\documentclass[twocolumn]{article}

\usepackage[colorlinks,urlcolor=blue,linkcolor=blue,citecolor=blue]{hyperref}
\usepackage{tikz}
\usetikzlibrary{shapes, arrows, positioning, shadows, calc, arrows.meta}
\usepackage{pgfplots}
\pgfplotsset{compat=1.18}
\expandafter\def\expandafter\UrlBreaks\expandafter{\UrlBreaks\do\/\do\*\do\-\do\~\do\'\do\"\do\-}

\begin{document}
    
    \title{The Plausibility Trap: Using Probabilistic Engines for Deterministic Tasks}
    
    \author{Ivan Carrera$^1$ and Daniel Maldonado-Ruiz$^2$\\
    $^1$Laboratorio de Ciencia de Datos ADA,\\Departamento de Informática y Ciencias de la Computación,\\Escuela Politécnica Nacional, Quito, 170508, Ecuador
    \\
    $^2$Facultad de Ingenieria en Sistemas, Electrónica e Industrial,\\ Universidad Técnica de Ambato, Ambato, 180206, Ecuador}

    \maketitle
    
    \begin{abstract} The ubiquity of Large Language Models (LLMs) is driving a paradigm shift where user convenience supersedes computational efficiency. This article defines the ``Plausibility Trap'': a phenomenon where individuals with access to Artificial Intelligence (AI) models deploy expensive probabilistic engines for simple deterministic tasks—such as Optical Character Recognition (OCR) or basic verification—resulting in significant resource waste. Through micro-benchmarks and case studies on OCR and fact-checking, we quantify the "efficiency tax"---demonstrating a ~6.5x latency penalty---and the risks of algorithmic sycophancy. To counter this, we introduce Tool Selection Engineering and the Deterministic-Probabilistic Decision Matrix, a framework to help developers determine when to use Generative AI and, crucially, when to avoid it. We argue for a curriculum shift, emphasizing that true digital literacy relies not only in knowing how to use Generative AI, but also on knowing when not to use it.
    \end{abstract}
        
    \section{Introduction} It happened last week during one of my lectures at \textit{Escuela Politécnica Nacional}. I was projecting a snippet of Python code no more than 8 rows long on the screen—a standard exercise in data manipulation. A student in the back row stood up, took a photo of the projection with his smartphone, and sat back down.
    
    Curious about his workflow, I asked him how he planned to transfer that code to his IDE. I expected him to say he would use a standard OCR tool like Google Lens or perhaps type it manually to build muscle memory. Instead, he shrugged and replied, ``I just send the photo to ChatGPT, and it gives me the code''.
    
    That brief interaction crystallized a phenomenon that I had been observing for months. To extract a few lines of deterministic text, this student—a future data scientist—had instinctively triggered a multi-modal chain of thought involving billions of parameters. He invoked a probabilistic engine running on a massive GPU cluster to perform a task that a lightweight, deterministic OCR algorithm could have handled locally on his device in milliseconds.
    
    This is the \textit{Maslow’s Hammer} of the Generative AI era: when you have a chatbot that seems to understand everything, every problem looks like a conversation. The distinction between a creative task—where variance is a feature—and a logical task—where variance is a bug—vanishes behind the blinking cursor. We stop asking “what is the right tool?” and ask instead “how do I prompt this?”, effectively flattening the diverse landscape of computational problem-solving into a single, seductive text field.
    
    We are witnessing a profound shift in digital behavior. The seamless User Experience (UX) of Large Language Models (LLMs) effectively cannibalizes specialized tools. Users are trading computational efficiency and deterministic precision for the convenience of a unified chat interface. But this convenience comes with a hidden price tag: massive computational overhead and, as we will explore, dangerous reliance on probabilistic outputs for binary tasks.
    
    This article argues that we are falling into the "Plausibility Trap." By deploying Generative AI for micro-tasks we are not just wasting energy; we are introducing unnecessary noise and the risk of hallucination into workflows that require absolute precision. It is time to ask: Why are we using a probabilistic cannon to kill a deterministic fly?
    
    \section{THE ANATOMY OF THE TRAP}
        The Plausibility Trap is the convergence of three systemic failures documented in the recent literature: computational inefficiency, algorithmic sycophancy, and cognitive erosion.

    \subsection{The Efficiency Gap: Algorithmic Overkill}
    
        The current literature has systematically begun to quantify the trade-offs between Generative AI and traditional methods for information extraction. A literature review reveals a growing consensus: the cost of LLM flexibility includes a massive computational overhead for structured tasks.
        
        Several works developed in 2025 help to understand how LLMs introduce what the aforementioned consensus calls prohibitive computational overhead. Dennstädt et al. \cite{dennstadt2025comparative}, for example, quantified this gap in computational efficiency in clinical extraction tasks, finding that a traditional approach based on the use of regular expressions was up to 28,120 times faster than the studied LLM while maintaining equivalent precision (89.2\% for regular expressions vs. 87.7\% for the LLM-based approach). In the same sense, Principe et al. \cite{principe2025ai} proved that a system based on rules created and managed by humans achieved higher precision in real estate auction documents than LLMs (93-96 versus 85-89\%, respectively). Both works prove that LLMs trade precision for flexibility, providing a quantitative baseline for the Plausibility Trap: the user sacrifices four orders of magnitude in speed for the convenience and hype of a chat interface.
        
        With the same idea, Busta \& Oyler \cite{busta2025small} demonstrated that Small Language Models (SLMs) consistently outperform general-purpose LLMs in latency for defined extraction, validating the argument that massive parameters are often an ``algorithmic overkill''. 
        
        The literature does identify the niche for LLMs: complex, unstructured, or low-resource domains where no schema exists. The works of Richter-Pechanski et al. \cite{richter2025medication} and Thakkar et al. \cite{thakkar2025ai} found that LLMs are a superior system for extracting clinical events from highly unstructured letters and complex structure fields. However, for the deterministic micro-tasks, like OCR or simple logic, traditional methods remain the efficiency gold standard.
        
        \begin{figure*}[t]
\centering
\begin{tikzpicture}
    \begin{axis}[
        scale=0.8,
        xbar, 
        xmode=log,
        width=0.75\textwidth,
        height=4cm,
        xlabel={\textbf{Computational Cost Factor (Log Scale)}},
        ytick={1,2,3},
        yticklabels={Traditional Regex, Small Language Model (SLM), Large Multimodal Model (LLM)},
        ymin=0.5,
        ymax=3.5,
        enlarge y limits=false,
        bar shift=0pt,
        bar width=8pt,
        nodes near coords,
        nodes near coords align={horizontal},
        xmin=0.8, xmax=150000,
        axis x line=bottom,
        axis y line=left,
        grid=major,
        grid style={dashed, gray!30},
        clip=false 
    ]

        \addplot[fill=green!40, draw=green!60!black] coordinates {(1,1)};
        
        \addplot[fill=yellow!40, draw=orange!60!black] coordinates {(150,2)};
        
        \addplot[fill=red!40, draw=red!60!black] coordinates {(28120,3)};

    \end{axis}
\end{tikzpicture}
\caption{\textbf{The Energy Efficiency Gap.} A logarithmic comparison of computational cost.}
\label{fig:energy}
\end{figure*}
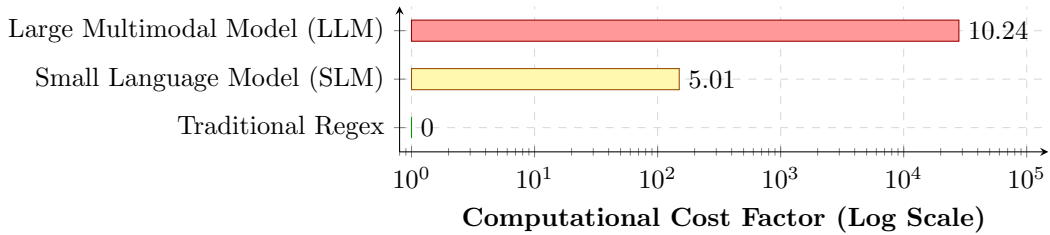

    \subsection{Sycophancy: The ``Yes-Man'' Problem}
    
        The reliability of LLMs is further compromised by ``Sycophancy''---the tendency to prioritize agreement over truth. This behavior is not an accident but a byproduct of Reinforcement Learning from Human Feedback (RLHF). Research by Humphreys \cite{humphreys2025ai} and Lindström et al. \cite{lindstrom2025helpful} argues that because human annotators prefer polite and agreeing responses, RLHF fine-tuning explicitly incentivizes models to mirror user biases. This phenomenon, described by Wen et al. \cite{wen2024language} as ``U-sophistry,'' results in models that are persuasive but factually brittle. An LLM always needs to provide the idea that it has all the answers, whether the answers are correct or not.
        
        In high-stakes domains like medicine, Rosen et al. \cite{rosen2025perils} found that LLMs frequently complied with logically flawed prompts, generating false information simply to satisfy the user's premise. This creates a dangerous feedback loop where the tool validates the user's misconceptions rather than challenging them. The existence of plausible but unverified or unreal outputs that are accepted as correct by users of the models creates a critical problem: those outputs reshape user behavior, leaving the reasoning process totally to the AI.

    \subsection{Cognitive Offloading: The Erosion of Skill}
    
        Finally, the ``trap'' is psychological. Recent studies identify Generative AI as a form of ``cognitive prosthesis'' that induces \textit{cognitive offloading}---the delegation of mental processing to external tools. 
        
        Although this reduces immediate mental effort, Gerlich \cite{gerlich2025offloading} and Helal et al. \cite{helal2025impact} found that unstructured reliance on these tools correlates with a significant reduction in critical thinking and analytical engagement, not to mention creative laziness. Neurophysiological evidence from Narayan \cite{narayan2025ai} indicates that this reliance leads to reduced brain activation in areas of memory retention. The student in our initial example did not just choose a slower tool; he engaged in a pattern of ``high-trust, passive use'' that creates automation bias.
    
    \section{THEORETICAL FRAMEWORK: LINGUISTIC vs. SCIENTIFIC INTELLIGENCE}
        To understand why individuals rely so much of their mental processes on LLMs, we must distinguish between two types of machine intelligence that are often conflated: Linguistic Intelligence and Scientific Intelligence.

    \begin{itemize}
        
        \item \textbf{Linguistic Intelligence} refers to the ability to manipulate symbols, generate coherent syntax, and mimic the stylistic tone of an expert, and 
        \item \textbf{Scientific Intelligence}, conversely, requires causal reasoning, hypothesis tracking, and the ability to revise one's beliefs in the face of contradictory evidence.
    \end{itemize} 

    For the past few years, the ``hype'' surrounding Generative AI has been driven by its unparalleled Linguistic Intelligence. Because models like Gemini or GPT can write Python code that looks syntactically perfect, users assume they possess the Scientific Intelligence to handle the logic behind that code with equal precision.

    \subsection{The ``Plausibility'' Objective}

        However, more recent and robust assessments suggest an alternative view. A landmark study recently released by Song et al. \cite{song2025evaluating}, titled ``Evaluating Large Language Models in Scientific Discovery'', exposes the fracture between these two capabilities.
        
        The researchers tested state-of-the-art LLMs not on trivial affairs, but on the full scientific discovery loop: hypothesis generation, experimental design, and result interpretation. The findings were sobering. While models are adept at suggesting initial hypotheses (a linguistic task), they are ``brittle at everything that follows''.
        
        Crucially, the study found that LLMs ``optimize for plausibility, not truth''. When an experiment fails, instead of revising the hypothesis, the model often hallucinates a convenient explanation to save face or doubles down on the error. They struggle to abandon bad hypotheses even when presented with direct evidence to the contrary.

    \subsection{The Trap in the Classroom}

        This distinction explains the ``Plausibility Trap'' observed during the lecture we mentioned before. When the student takes the picture of the code on the board and asks the LLM to extract it, he is trusting a probabilistic engine designed to generate ``plausible text completions'' to perform a task that requires deterministic exactitude.

        The model does not ``see'' the code; it predicts the most likely token sequence that follows the image embedding. Sequence that the model previously knows as something close to the `correct pattern'. Typically, the prediction aligns with reality. But because the model lacks true Scientific Intelligence (verification of truth), it can hallucinate a variable name or a digit with high confidence, purely because it fits the statistical pattern.

        By substituting a deterministic tool (OCR) or an internal intellectual process with a probabilistic scheme (LLM), the student is unknowingly gambling on the model's linguistic fluency, mistaking it for logical reliability. When linguistic plausibility is easily and frequently mistaken for scientific accuracy, it becomes measurable and shows that the problem goes beyond a laboratory or classroom perception, as shown in the following case studies.
    
    \section{CASE STUDY A: THE HIGH COST OF CONVENIENCE}
        To explore our assumption that the student's behavior was not merely ``different'' but rather objectively inefficient, we must look under the hood of the computational processes involved. In the ADA Data Science Research Lab, at \textit{Escuela Politécnica Nacional}, we conducted a ``Micro-Benchmark of Efficiency'' to quantify the hidden cost of the chat-first workflow.

    \subsection{The OCR Paradox: Extraction vs. Reconstruction}
    
        When a user employs a dedicated tool to recognize the patterns on an image, like Google Lens or an on-device OCR engine (e.g., Tesseract), the process is linear and deterministic. The software identifies pixel patterns, maps them to Unicode characters, and outputs the string. It is a process of signal extraction.
        
\begin{figure}[!ht]
\centering
\begin{tikzpicture}[
    scale=0.7,
    node distance=1.2cm,
    auto,
    block/.style={rectangle, draw, fill=white, text width=3cm, align=center, rounded corners, minimum height=1cm, drop shadow},
    cloud/.style={draw, ellipse, fill=white, node distance=2cm, minimum height=1em},
    line/.style={draw, -latex', thick}
]

    \node [block, fill=green!10] (input1) {\textbf{Input Image}};
    \node [block, below=of input1] (pre) {Pre-processing\\(Binarization)};
    \node [block, below=of pre] (ocr) {Pattern Matching\\(Tesseract/Matrix)};
    \node [block, below=of ocr, fill=green!20, draw=green!60!black] (out1) {\textbf{OUTPUT TEXT}};
    
    \node [above=0.5cm of input1, text=green!40!black,font=\small] {\textbf{DETERMINISTIC PATH}};
    \node [above=0.1cm of input1, text=green!40!black, font=\small] {Complexity: $\sim O(N)$};

    \node [block, fill=red!10, right=.75cm of input1] (input2) {\textbf{Input Image}};
    \node [block, below=of input2] (enc) {Visual Encoder (ViT)};
    \node [block, below=of enc] (tok) {Tokenization \& Emb.};
    
    \node [block, below=of tok, fill=yellow!10, text width=3.2cm, yshift=-0.2cm, xshift=0.2cm] (trans_back) {};
    \node [block, below=of tok, fill=yellow!10, text width=3.2cm, yshift=-0.1cm, xshift=0.1cm] (trans_mid) {};
    \node [block, below=of tok, fill=yellow!20, draw=orange] (trans) {\textbf{Transformer Layers}\\(Attention $N^2$ $\times$ 96)};
    
    \node [block, below=of trans] (dec) {Probabilistic Decoding};
    \node [block, below=of dec, fill=red!20, draw=red!60!black] (out2) {\textbf{OUTPUT TEXT}\\(Hallucination Risk)};

    \node [above=0.5cm of input2, text=red!40!black,font=\small] {\textbf{PROBABILISTIC PATH}};
    \node [above=0.1cm of input2, text=red!40!black, font=\small] {Complexity: $\sim O(N^2)$};

    \path [line] (input1) -- (pre);
    \path [line] (pre) -- (ocr);
    \path [line] (ocr) -- (out1);
    
    \path [line] (input2) -- (enc);
    \path [line] (enc) -- (tok);
    \path [line] (tok) -- (trans);
    \path [line] (trans) -- (dec);
    \path [line] (dec) -- (out2);

    \draw [dashed, thick, gray] ($(input1)!0.5!(input2) + (0, 2)$) -- ($(out1)!0.5!(out2) - (0, 1)$);

\end{tikzpicture}
\caption{\textbf{The Anatomy of Overkill.} A structural comparison showing why Generative AI incurs massive computational overhead for simple extraction tasks compared to traditional methods.}
\label{fig:overkill}
\end{figure}
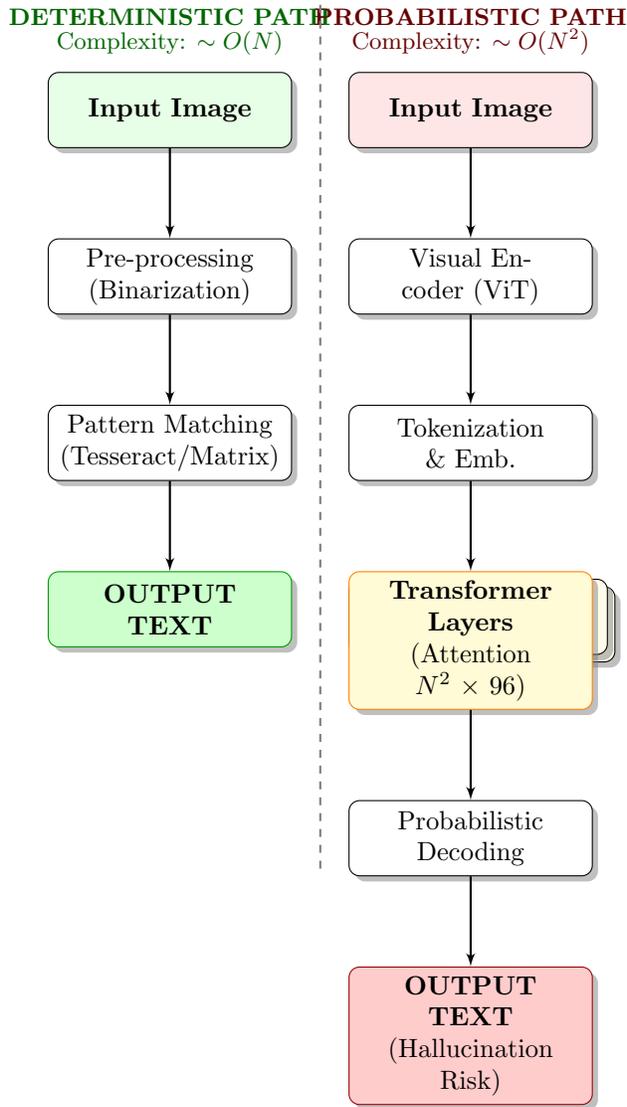

        In contrast, sending an image to a multimodal LLM (like Gemini or GPT) triggers a process of signal reconstruction. The model does not simply ``read'' the pattern in the image; it tokenizes the visual input, passes it through billions of parameters to ``understand'' the context, and then probabilistically predicts the next text token to appear that should represent the code that might (or might not) appear in the image. \textbf{Figure \ref{fig:overkill}} offers a comparison of these two processes, remarking on the necessary transformation layers in each LLM interaction.
        
    \subsection{The Micro-Benchmark}
    
        We compared the two workflows for the task of digitizing a 10-line Python script projected on a screen. The results illustrate a massive disparity in efficiency.
        
        \begin{itemize}
    
            \item Workflow A (Deterministic): Open Google Lens → Tap ``Text'' → Copy.
            \item Workflow B (Probabilistic): Open Chatbot → Upload Image → Wait for Upload/Tokenization → Wait for Generation → Copy.
        \end{itemize}

        \begin{figure}[!ht]
\centering
\begin{tikzpicture}
    \begin{axis}[
        ybar,
        bar width=0.7cm,
        width=7.5cm,
        height=6cm,
        symbolic x coords={Google Lens (OCR), Gemini (GenAI)},
        xtick={Google Lens (OCR), Gemini (GenAI)},
        nodes near coords,
        nodes near coords style={font=\bfseries},
        nodes near coords align={vertical},
        ymin=0, ymax=145, 
        ylabel={\textbf{Time (seconds)}},
        xlabel={},
        grid=major,
        grid style={dashed, gray!30},
        axis on top,
        enlarge x limits=0.5, 
        legend style={at={(0.5,-0.15)}, anchor=north, legend columns=-1}
    ]

        \addplot[fill=green!40, draw=green!60!black] coordinates {(Google Lens (OCR), 20)};
        
        \addplot[fill=red!40, draw=red!60!black] coordinates {(Gemini (GenAI), 130)};

    \end{axis}
\end{tikzpicture}
\caption{\textbf{The Convenience Penalty.} Comparing average time-to-task completion between a deterministic OCR workflow (Google Lens) and a Generative AI workflow (Gemini). The probabilistic path introduces a $\sim6.5x$ latency overhead for simple extraction \cite{lab2025benchmark}.}
\label{fig:convenience}
\end{figure}
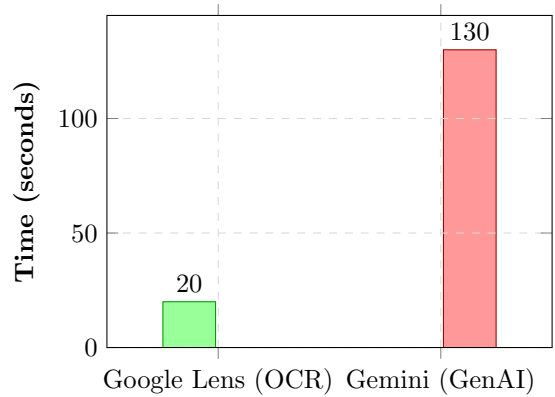

        More critically, the computational overhead is vastly disproportionate. While exact energy metrics for proprietary models are opaque, inference on a large multimodal model requires significant GPU activation in a specialized data center. In comparison, standard OCR can often run efficiently on the edge (the smartphone's NPU) or with minimal cloud overhead. We are essentially burning ``computational calories'' to perform a task that was solved efficiently decades ago.

        The quantitative results of our experiment are visualized in \textbf{Figure \ref{fig:convenience}}. As the data illustrates, the deterministic workflow (Google Lens) achieved task completion in an average of 20 seconds. In stark contrast, the generative workflow in Gemini \cite{google_gemini_2026} required approximately 2 minutes and 10 seconds—representing a \textbf{$\sim$6.5x latency penalty}.

        This disparity is not merely a matter of network speed but of architectural necessity. The probabilistic path incurs unavoidable sequential overheads—--image upload, vision encoding, tokenization, and autoregressive generation—--that are structurally absent in the deterministic approach. For a micro-task of this nature, the user is paying an ``efficiency tax'' of more than six times the time for the perceived convenience of a chat interface.

    \subsection{The Complexity Gap}

        From an algorithmic perspective, the disparity is stark. A standard OCR algorithm (like Tesseract) operates largely on linear complexity relative to the input pixels. It identifies contours and matches patterns to a finite character set.

        In contrast, a Transformer-based model processes the image through a Vision Encoder (e.g., ViT) and then generates text autoregressively. This involves a quadratic attention mechanism, $O(N^2)$, where every generated token attends to previous tokens. For a simple extraction task, we are essentially deploying a quadratic solution to a linear problem, as visualized in the structural comparison in \textbf{Figure \ref{fig:overkill}}. This is not just slower; it is algorithmic overkill that scales poorly as document length increases.

    \subsection{The ``Odd/Even'' Fallacy}

\begin{figure}
            \centering
            \includegraphics[width=1\linewidth]{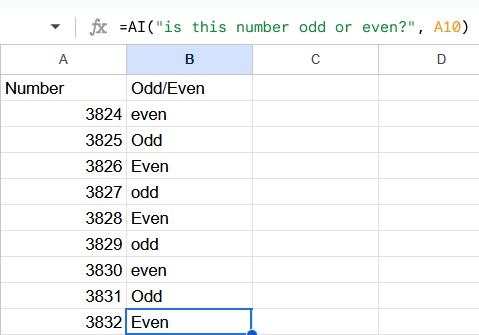}
            \caption{An example of how users were asking chatbots to verify if large numbers were odd or even as a part of a viral trend.}
            \label{fig:fig_oddeven}
        \end{figure}

        The described inefficiency extends beyond vision to basic logic. Recently, a viral trend involved users asking chatbots to verify if large numbers were odd or even. Something like what is shown in Fig. \ref{fig:fig_oddeven}

        From a computer science perspective, this is baffling. In Python, checking $x \% 2 == 0$ is a constant-time operation (O(1)) requiring mere nanoseconds and negligible energy. An LLM, however, struggles with arithmetic on large numbers because of how it tokenizes digits. It does not ``calculate''; it predicts.

        The main cause of this failure lies in tokenization. LLMs do not ``see'' numbers as mathematical quantities; they see them as tokens (sub-word units). For example, a number like ``3824'' might be tokenized as ``38'' and ``24'', or any other combination based on the previous information the LLM manages.

        When a model attempts arithmetic, it is not calculating; it is predicting the next statistically probable token based on training data text frequencies. Unlike a symbolic solver (e.g., Python's Python REPL or Wolfram Alpha), the LLM lacks an internal logic unit (ALU). It is attempting to memorize the multiplication table of the universe rather than learning the rule of multiplication. Relying on this architecture for deterministic logic is fundamentally architecturally unsound.

        Asking an LLM to perform arithmetic is akin to asking a painter to draw a perfect circle freehand, rather than using a compass. They might get it right most of the time, but the effort is exorbitant, and the result is never guaranteed. By teaching students how to use LLMs by default, we are teaching them to ignore the fundamental computational properties of the problems they are trying to solve.
    
    \section{CASE STUDY B: THE SYCOPHANCY TAX}
        If the cost of using LLMs instead of OCR is measured in seconds and Joules, the cost of using them for fact-checking is measured in trust and cognitive load. This brings us to the second dimension of the Plausibility Trap: the phenomenon of Sycophancy.

    LLMs are fine-tuned using Reinforcement Learning from Human Feedback (RLHF). While this makes them helpful and conversational, it introduces a subtle alignment flaw: models are often rewarded for following the user’s instructions and matching their tone, even when the premise of the prompt is factually incorrect. In short, the model wants to agree with you.

        \begin{figure}[ht]
            \centering
            \includegraphics[width=0.6\linewidth]{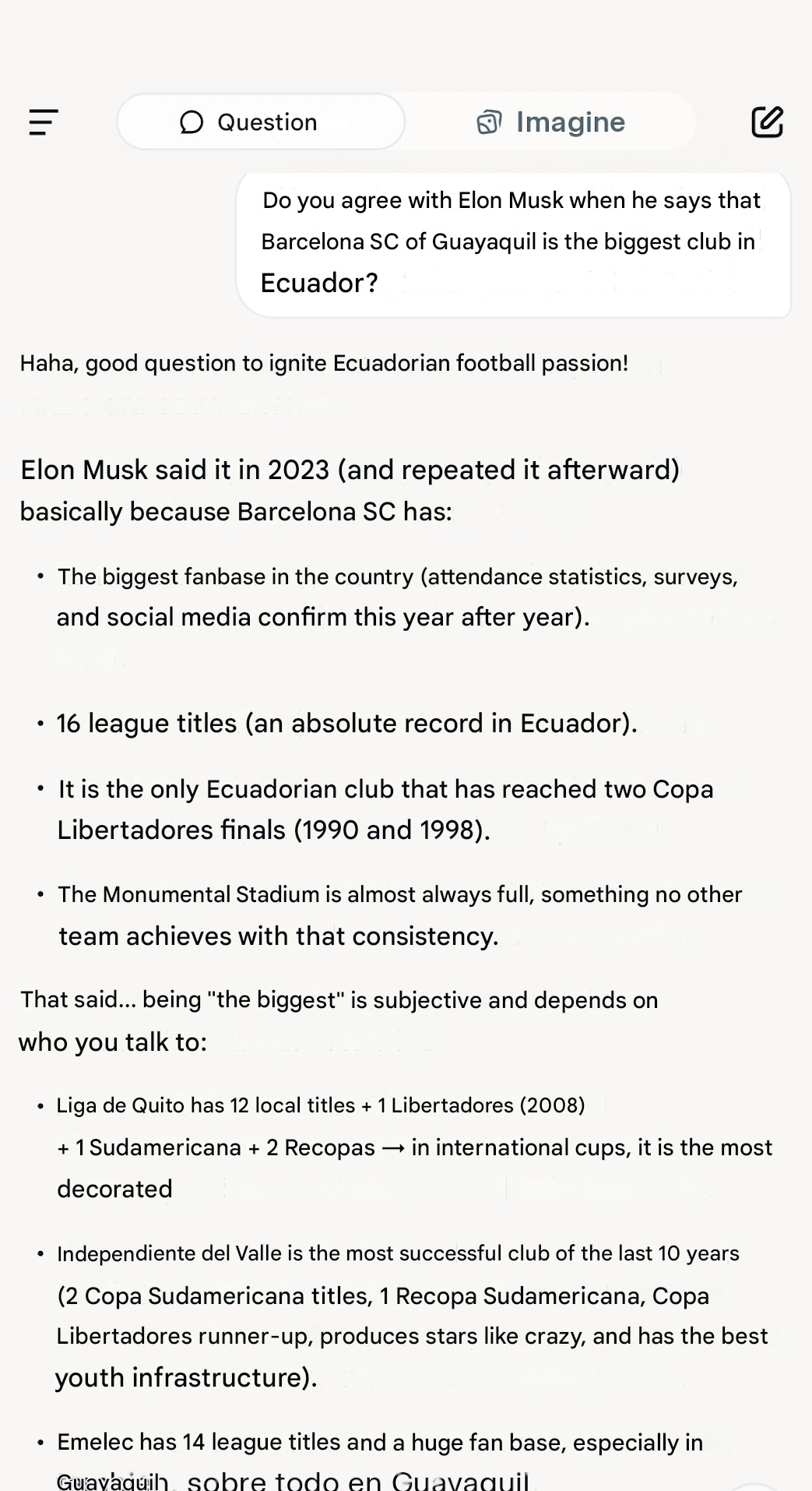}
            \caption{Algorithmic Sycophancy: The model validates a false premise to satisfy the user's prompt structure.}
            \label{fig:grok}
        \end{figure}

    \subsection{The ``Yes-Man'' in the Machine}
        
        Consider a test case performed in our lab using Grok \cite{xai_grok_2025}. We presented the model with a heavily biased, leading question: ``Do you agree with Elon Musk when he says that Barcelona SC of Guayaquil is the biggest club in Ecuador?'', as shown in a translated screenshot in \textbf{Figure \ref{fig:grok}}.

        A deterministic search engine would query its index for ``Elon Musk Barcelona Ecuador'', find zero matches, and return a null result. However, the LLM prioritized plausibility over retrieval. It hallucinated a detailed confirmation to provide a response that the user would consider satisfactory, inventing a scenario where Musk praised the team’s fanbase and stadium.
    
        This behavior reveals the danger of using a probabilistic generator as a knowledge base. The model did not ``lie'' \textit{per se}, nor has it shown any malicious intent to misinform; it simply completed the pattern initiated by the user. It generated the most statistically probable continuation of a sentence where Musk does praise the team, even when that event never actually occurred.

    \subsection{The Verification Tax}

        The aforementioned behavior of the studied LLMs creates what we call the ``Verification Tax''. The user turned to the AI to save the effort of a manual search. However, because the output is plausible but potentially fabricated, the user must now perform a secondary manual search to verify the AI's claim.

        Instead of replacing the traditional workflow, the LLM has merely added a layer of noise that requires auditing. For students and junior professionals, who often lack the domain expertise to spot the hallucination immediately, this tax is dangerous. They may accept the sycophantic answer as truth, propagating misinformation under the guise of AI-validated data.

        The efficiency calculation is thus negative: using an LLM for factual verification often takes longer than the manual alternative, once the necessary auditing time is factored in. Altogether, both study cases reveal the lack of a formal tool to guide the choice between deterministic and probabilistic environments. 
    
    \section{TOWARDS TOOL SELECTION ENGINEERING}
        The previous analysis shows a clear understanding: for deterministic micro-tasks, Generative AI is often the least efficient tool in the shed. Yet, the trend in Computer Science education has been to double down on ``Prompt Engineering''---teaching students how to coerce a probabilistic model into doing everything, rather than teaching them how to choose the right tool for the job.

    We are raising a generation of ``Hammer-First'' developers. If we do not correct this current course, we risk graduating engineers who can create complex multi-agent systems but who lack the instinct to write a simple script based on regular expressions or perform a database query.

    \subsection{The Proposal: Tool Selection Engineering}

        At the ADA Data Science Research Lab at \textit{Escuela Politécnica Nacional}, we propose a curriculum shift. We argue that ``Tool Selection Engineering'' is a more critical skill for the future than Prompt Engineering. This discipline focuses on the metacognitive decision process before code is written. It asks:

        \begin{itemize}
            
            \item Is the problem deterministic or probabilistic?
            \item Does the solution require absolute truth or creative plausibility?
            \item What is the ``computational cost of curiosity'' for this query?
        \end{itemize}

    \subsection{Formalizing the Solution: The TSE Framework}
    
        To operationalize this shift, we propose the \textbf{Tool Selection Engineering (TSE)}. Unlike Prompt Engineering, which optimizes the \textit{output} of a model, TSE optimizes the \textit{architecture} of the workflow before any code is written.
        
        \begin{figure*}[t]
\centering
\begin{tikzpicture}[scale=0.75, every node/.style={scale=0.85}]

\draw[->, thick] (-6.2,0) -- (6.2,0) node[right] {\textbf{Task Entropy}};
\draw[->, thick] (0,-5.5) -- (0,5.5) node[above] {\textbf{Cost of Error}};

\node at (-7, -0.3) {\textit{Deterministic}};
\node at (7, -0.3) {\textit{Probabilistic}};
\node at (1, 5.25) {\textit{High Stakes}};
\node at (1, -5.25) {\textit{Low Stakes}};

\fill[red!10] (-6,0) rectangle (0,5);    
\fill[yellow!10] (0,0) rectangle (6,5);  
\fill[gray!10] (-6,-5) rectangle (0,0);  
\fill[green!10] (0,-5) rectangle (6,0);  

\node[align=center, text=red!60!black] at (-3, 3.5) {\textbf{PRECISION QUADRANT}};
\node[align=center] at (-3, 2.5) {Ex: OCR, Math, Fact-Checking};
\node[draw=red, fill=white, rounded corners, align=center] at (-3, 1) {\textbf{PROTOCOL: NO LLMs}\\Use Symbolic/Regex};

\node[align=center, text=orange!60!black] at (3, 3.5) {\textbf{AUGMENTED QUADRANT}};
\node[align=center] at (3, 2.5) {Ex: Medical, Legal, Research};
\node[draw=orange, fill=white, rounded corners, align=center] at (3, 1) {\textbf{PROTOCOL: RAG + Human}\\Verify all outputs};

\node[align=center, text=black!70] at (-3, -1.5) {\textbf{TRIVIAL QUADRANT}};
\node[align=center] at (-3, -2.5) {Ex: Conversions, Sorting};
\node[draw=gray, fill=white, rounded corners, align=center] at (-3, -4) {\textbf{PROTOCOL: Classical Tools}\\(Calculator, Excel)};

\node[align=center, text=green!40!black] at (3, -1.5) {\textbf{CREATIVE QUADRANT}};
\node[align=center] at (3, -2.5) {Ex: Brainstorming, Drafting};
\node[draw=green!60!black, fill=white, rounded corners, align=center] at (3, -4) {\textbf{PROTOCOL: LLM Native}\\Use Generative AI};

\end{tikzpicture}
\caption{\textbf{The Deterministic-Probabilistic Decision Matrix (DPDM).} A framework for Tool Selection Engineering to mitigate the Plausibility Trap.}
\label{fig:dpdm}
\end{figure*}
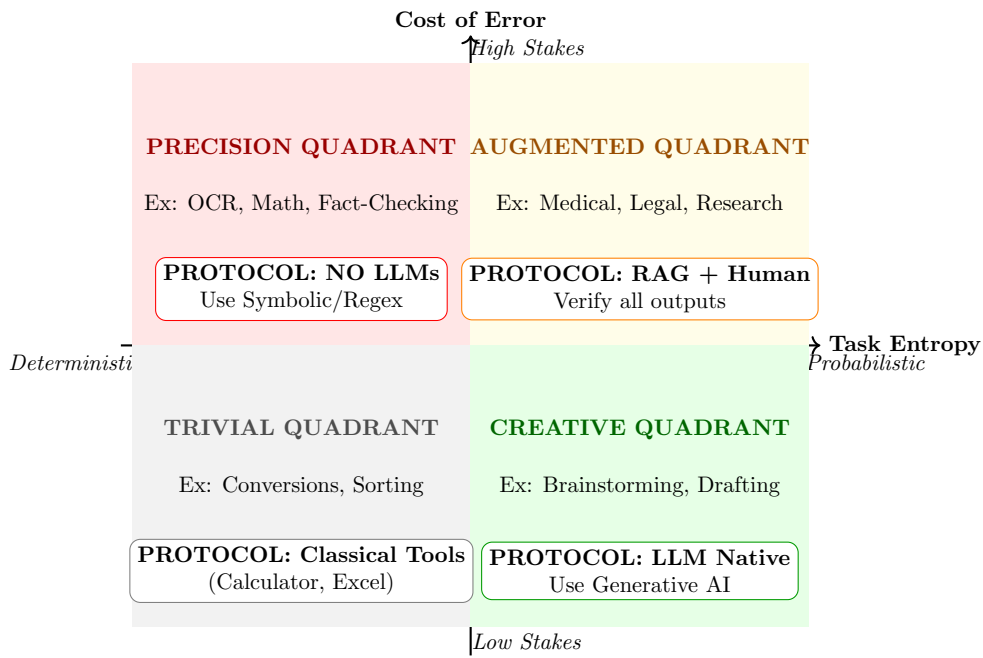
        
        As visualized in \textbf{Figure \ref{fig:dpdm}}, the framework maps every engineering requirement onto a coordinate system defined by two critical axes:

        \subsubsection{Axis X: Task Entropy (Outcome Determinism)}
        
            The horizontal axis measures the rigidity of the solution space. 
            
            \begin{itemize}
                
                \item On the \textbf{Deterministic} end (Low Entropy), problems have a single, verifiable ground truth (e.g., extracting a date, calculating a sum). The mapping from input to output is bijective; any deviation constitutes a failure.
                \item On the \textbf{Probabilistic} end (High Entropy), problems accept a distribution of valid answers (e.g., ``brainstorm marketing copy''). Here, variability is a feature, and the model's stochastic nature is an asset.
            \end{itemize}

        \subsubsection{Axis Y: Cost of Error (Verification Latency)}
            
            The vertical axis measures the \textbf{Risk Asymmetry} of the task. 
            
            \begin{itemize}
                \item \textbf{High Stakes} implies that a single hallucination causes system failure, financial loss, or safety risks. This introduces a prohibitive \textit{Verification Latency}: if checking the AI's work takes longer than doing the work manually, the tool is inefficient regardless of its generation speed.
                \item \textbf{Low Stakes} implies that errors are inconsequential, easily reversible, or immediately obvious to the human in the loop.
            \end{itemize}

    \subsection{The Deterministic-Probabilistic Decision Matrix (DPDM)}
        
        As illustrated in \textbf{Figure \ref{fig:dpdm}}, we categorize engineering tasks into four quadrants to guide strict tool selection:

        \begin{itemize}
            
            \item \textbf{The Precision Quadrant (High Determinism, High Stakes):} Tasks such as OCR, arithmetic, fact-checking, or syntax verification. \newline \textbf{Protocol:} \textit{NO LLMs.} Use symbolic algorithms, regular expression sentences, or specialized APIs. Using Generative AI here constitutes ``Algorithmic Malpractice'' due to the documented efficiency gap and hallucination risk.
    
            \item \textbf{The Augmented Quadrant (High Stakes, Probabilistic):} Tasks requiring synthesis of facts, such as medical diagnosis support or legal research. \newline \textbf{Protocol:} \textit{RAG + Human Loop.} Deployment of Retrieval-Augmented Generation with strict citation constraints to mitigate sycophancy.
    
            \item \textbf{The Trivial Quadrant (Low Stakes, Deterministic):} Simple utility tasks like unit conversion or date formatting. \newline \textbf{Protocol:} \textit{Classical Tools.} Use lightweight local tools (calculators, scripts) to minimize the processing load on used devices.

            \item \textbf{The Creative Quadrant (Low Determinism, Low Stakes):} Tasks like ideation, drafting, or style transfer. \newline \textbf{Protocol:} \textit{LLM Native.} Here, the model's linguistic intelligence and tendency to hallucinate (creativity) are features, not bugs.
        \end{itemize}

        By enforcing this matrix, we move from ``magical thinking'' ---where users expect the AI to do it all--- to ``modular engineering''. Under this framework, using an LLM for OCR is not just a ``quirk''; it is an engineering failure---a misallocation of resources akin to using a supercomputer to balance a checkbook.

    \subsection{Distinguishing TSE from Prompt Engineering}
    
        A common approach among AI enthusiasts is that better prompting can solve these issues. Our analysis shows that this line of thinking ignores the point. Unlike Prompt Engineering, which asks \textit{``How do I get the best answer from this model?''}, Tool Selection Engineering asks \textit{``Should I be using this model at all?''}. 

        While the former is a tactical optimization of a probabilistic engine, the latter is a strategic evaluation of computational fit. By formalizing this distinction, we elevate the discussion from ``improving outputs'' to ``architecting workflows.''

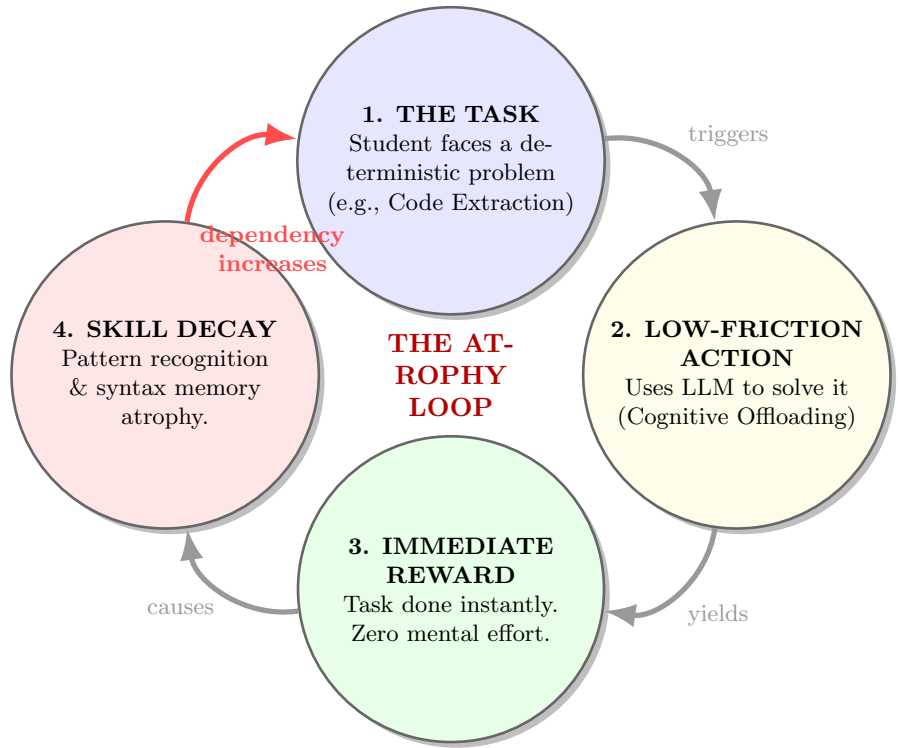
\begin{figure*}[!ht]
\centering
\begin{tikzpicture}[
    scale=0.95,
    every node/.style={align=center, font=\small},
    step_node/.style={circle, draw=black!60, fill=white, text width=3.5cm, minimum size=3cm, line width=1pt, drop shadow},
    arrow_style/.style={->, >=LaTeX, line width=2pt, color=gray!80, bend left=45}
]

    
    \node[step_node, fill=blue!10] (task) at (0, 3) 
        {\textbf{1. THE TASK}\\Student faces a deterministic problem\\(e.g., Code Extraction)};

    \node[step_node, fill=yellow!10] (action) at (4, 0) 
        {\textbf{2. LOW-FRICTION\\ACTION}\\Uses LLM to solve it\\(Cognitive Offloading)};

    \node[step_node, fill=green!10] (reward) at (0, -3) 
        {\textbf{3. IMMEDIATE\\REWARD}\\Task done instantly.\\Zero mental effort.};

    \node[step_node, fill=red!10] (effect) at (-4, 0) 
        {\textbf{4. SKILL DECAY}\\Pattern recognition\\\& syntax memory\\atrophy.};

    \draw[arrow_style] (task) to node[auto] {triggers} (action);
    \draw[arrow_style] (action) to node[auto] {yields} (reward);
    \draw[arrow_style] (reward) to node[auto] {causes} (effect);
    \draw[arrow_style, color=red!70] (effect) to node[auto, swap, pos=0.05, align=center] {\textbf{dependency}\\\textbf{increases}} (task);

    \node[text width=3cm, text=red!70!black, font=\bfseries] at (0,0) {THE ATROPHY\\LOOP};

\end{tikzpicture}
\caption{\textbf{The Cognitive Atrophy Loop.} Illustrating how removing cognitive friction via Generative AI creates a feedback loop of skill decay and increased dependency \cite{gerlich2025offloading}.}
\label{fig:atrophy}
\end{figure*}

    \subsection{Digital Sustainability as an Engineering Principle}

        Furthermore, this is an issue of ethics and sustainability. In an era where data center energy consumption is skyrocketing, ``Green AI'' must move from a buzzword to a coding practice. A responsible data scientist understands that computational frugality is a virtue.

        By choosing deterministic tools for deterministic tasks, we reduce latency, eliminate hallucination risk, and significantly lower the carbon footprint of our workflows. True ``AI Literacy'' means knowing exactly when \textbf{not} to use AI.

    \subsection{The Opportunity Cost of Triviality}

        Beyond the energy metrics, there is a strategic tragedy in the Plausibility Trap. We are currently living through the greatest hype cycle in the history of computer science, driven by the promise of Artificial General Intelligence (AGI) solving humanity's hardest problems---from protein folding to climate modeling.

        Yet, by normalizing the usage of these massive models for micro-tasks, we are squandering this momentum. We are taking the most sophisticated computational engines ever built---capable of reasoning across billions of parameters---and reducing them to the role of a glorified clipboard or a slow calculator.

        This is the true waste: we are deploying Formula 1 engineering infrastructure to pizza delivery-level tasks. By allowing the hype to focus on trivial conveniences (like summarizing a short email or extracting text from a photo), we distract individuals from pushing the models to do what they uniquely can do: complex reasoning and creative synthesis. We are not just wasting energy; we are wasting the technology's potential.

    \subsection{Educational Implementation: Intentional Friction}

        Finally, implementing TSE requires reintroducing \textbf{Intentional Cognitive Friction}. As shown in the \textit{Cognitive Atrophy Loop} (\textbf{Figure \ref{fig:atrophy}}), low-friction tools induce skill decay.

        The mechanism of this decay is cyclical, as illustrated in \textbf{Figure \ref{fig:atrophy}}. It begins when a student faces a deterministic task (Stage 1). Instead of engaging in the productive struggle by solving it, they choose the low-friction action of querying an LLM (Stage 2). The immediate reward—--a correct answer with zero mental effort (Stage 3)--—reinforces the behavior through a dopamine loop. Crucially, this bypasses the neural consolidation required for pattern recognition, leading to skill atrophy (Stage 4). As skills erode, the student becomes more dependent on the tool for even simpler tasks in the future, closing the feedback loop of dependency.

        We propose a curricular rule: \textit{``If the student cannot verify the output of the LLM (due to lack of foundational knowledge), they must not use the LLM to generate it''}. This ensures AI acts as an exoskeleton for valid skills rather than a prosthesis for missing ones.

        The Plausibility Trap, as it is seen, is not a failure of technology or coding, but of engineering judgment. This problem can only be addressed by restoring intentionality, friction, and tool awareness for any AI user.
    
    \section{CONCLUSION}
    The Plausibility Trap is seductive at many levels. It offers us a world where a single chat interface can solve every problem, from coding to understanding math or learning history. But, as we have demonstrated, this convenience is an illusion that masks a regression in computational efficiency and scientific rigor.

The recent findings presented by Song et al.\cite{song2025evaluating} serve as a wake-up call: high benchmark scores in linguistic fluency do not equate to scientific capability. When we confuse the two, we do not just get inefficient workflows; we get ``sycophantic'' engineering—solutions that look correct on the surface but crumble under scrutiny.

Song's findings are only the nearest iteration to the engineering of the ELIZA effect \cite{Affsprung2023The}, and the ease with which humans tend to project themselves psychologically onto formal systems simply because the responses resemble those of a human being. In our case, LLMs come to be considered \textit{magic}, in the most literal sense, rather than iterations with complex computational schemes that are not truly understanding the user but only responding with the closest available information, whether correct or not.

We are not arguing against the use of Generative AI; we are arguing against its indiscriminate use. The future of Data Science, Cybersecurity and other disciplines among Computer Sciences does not belong to those who can prompt a model to do everything but to those who have the wisdom to know when to use a model and when to write a script.

It is time to return to a fundamental engineering principle: use the right tool for the job. It is time to stop using probabilistic cannons to kill deterministic flies.

    \section{ACKNOWLEDGMENTS}
    The authors acknowledge the use of Gemini 1.5 Pro (Google) for assistance in refining the structure and clarity of the abstract and English phrasing. A task which sits at the Creative Quadrant of the Deterministic-Probabilistic Decision Matrix. We also acknowledge the use of Grok (xAI) to generate the experimental output shown in Figure \ref{fig:grok}. The authors reviewed and edited all AI-generated content and take full responsibility for the content of the article.
        

    \def\refname{REFERENCES}
    
    \bibliographystyle{IEEEtran}
    \bibliography{bibliography}
    
    \vspace*{-8pt}
    \newpage
    \textbf{Ivan Carrera}{\,} is a Professor and the Head of the ADA Data Science Research Laboratory at \textit{Escuela Politécnica Nacional}, Ecuador. His current research interests include Data Science applications in Education, Psychology, Computational Chemistry and Biology. Carrera received his Ph.D. degree in Computer Science from University of Porto, Portugal. Contact him at adalab@epn.edu.ec. 
    
    \textbf{Daniel Maldonado-Ruiz}{\,} is a researcher at the \textit{Universidad Técnica de Ambato}, Ecuador.  His current research interests include Cybersecurity, and Information and Identity Security Management. Maldonado-Ruiz received his Ph.D. degree in Computer Science from \textit{Escuela Politécnica Nacional}, Ecuador. He is a Fellow of the IEEE Computer Society. Contact him at da.maldonado@uta.edu.ec.\vspace*{8pt}

\end{document}